\documentclass[10pt,journal,compsoc]{IEEEtran}
\usepackage{amsfonts,amsmath,amssymb,amsthm,bm}
\usepackage{graphicx,psfrag,epsf}
\usepackage{enumerate}
\usepackage{adjustbox}
\usepackage{dsfont}
\usepackage{comment}
\usepackage{xcolor}

\newcommand{\Real}{\mathbb{R}}

\newcommand{\E}{\mathbb{E}}

\ifCLASSOPTIONcompsoc
  \usepackage[nocompress]{cite}
\else
  \usepackage{cite}
\fi

\ifCLASSINFOpdf

\else
 
\fi

\begin{document}

\title{A Simple Spectral Failure Mode \\for Graph Convolutional Networks}

\author{Carey E.~Priebe,~Cencheng~Shen,~Ningyuan~(Teresa)~Huang,~Tianyi~Chen
\IEEEcompsocitemizethanks{\IEEEcompsocthanksitem Carey E.Priebe
is with
the Department of Applied Mathematics and Statistics (AMS), the Center for Imaging Science (CIS), and the Mathematical Institute for Data Science (MINDS), Johns Hopkins University. E-mail: cep@jhu.edu 
\IEEEcompsocthanksitem Cencheng Shen is with the Department of Applied Economics and Statistics, University of Delaware. E-mail: shenc@udel.edu 
\IEEEcompsocthanksitem Ningyuan (Teresa) Huang and Tianyi Chen are with the Department of Applied Mathematics and Statistics, Johns Hopkins University. E-mail: nhuang19@jhu.edu, tchen94@jhu.edu \protect
}
\thanks{This work was supported in part by the Defense Advanced Research Projects Agency under the D3M program administered through contract FA8750-17-2-0112,
the National Science Foundation HDR TRIPODS 1934979, the National Science Foundation DMS-2113099, the University of Delaware Data Science Institute Seed Funding Grant,
and by funding from Microsoft Research.
The authors thank Wade Shen for providing the motivation for this investigation, and also thank the editor and reviewers for valuable comments and suggestions that significantly improved the exposition of the paper.
}}

\markboth{IEEE Transactions on Pattern Analysis and Machine Intelligence, accepted, 2021}%
{Shell \MakeLowercase{\textit{et al.}}: Bare Demo of IEEEtran.cls for Computer Society Journals}

\IEEEtitleabstractindextext{%
\begin{abstract} 
Neural networks have achieved remarkable successes in machine learning tasks. This has recently been extended to graph learning using neural networks. However, there is limited theoretical work in understanding how and when they perform well, especially relative to established statistical learning techniques such as spectral embedding. In this short paper, we present a simple generative model where unsupervised graph convolutional network fails, while the adjacency spectral embedding succeeds. Specifically, unsupervised graph convolutional network is unable to look beyond the first eigenvector in certain approximately regular graphs, thus missing inference signals in non-leading eigenvectors. The phenomenon is demonstrated by visual illustrations and comprehensive simulations.
\end{abstract}



}

\maketitle

\IEEEdisplaynontitleabstractindextext

%
\IEEEpeerreviewmaketitle

\IEEEraisesectionheading{\section{Introduction}\label{sec:introduction}}

\IEEEPARstart{G}{raph} embedding aims to learn a low-dimensional representation for each node in a given graph, on which subsequent inferences can be directly performed. Recently, graph neural networks have emerged an attractive graph embedding solution due to the representation power: they can learn nonlinear mappings from the graph space to the embedding space, while traditional methods such as matrix factorization are restricted to linear mappings \cite{goyal2018graph}. On the other hand, powerful graph neural networks can fail if their architectures do not align with the downstream tasks \cite{Xu2020What}. 

The natural question is how well graph neural networks perform: Are they always the preferred graph embedding solution? And if not, can we characterize when and how they are sub-optimal? To answer the question, we present a simple and intuitive random graph generative model, where unsupervised graph convolutional network (GCN) fails while adjacency spectral embedding (ASE) -- a matrix factorization method -- succeeds. 

Specifically, we propose a latent position graph model where the classification information lies on the second eigenvector and perpendicular to the first eigenvector. The graph is approximately regular, and unsupervised GCN embedding mainly recovers the leading eigenvector of the graph adjacency matrix. Namely, unsupervised GCN can be sub-optimal for an approximately regular graph where the signal of the downstream task lies on non-leading eigenvectors, while ASE can succeed under proper dimension choice.

To our best knowledge, our results provide the first geometry characterization of GCN limitation under random graph models, which is an important aspect of unsupervised graph convolutional network that requires further investigation and improvement. In the following, we first introduce the necessary background, followed by the failure geometry, and extensive simulations to demonstrate the failure mode. 

\section{Background}

\subsection{Classification with Empirical Risk Minimization}

The classical statistical formulation of the classification problem consists of
$$(X,Y), (X_1,Y_1), \cdots, (X_m,Y_m) \stackrel{iid}{\sim} F_{XY}$$
where $\{(X_i,Y_i)\}_{i \in \{1,\cdots,m\}}$
is the training data, 
and $(X,Y)$ represents the to-be-classified test observation $X$
with true-but-unobserved class label $Y$.
We consider the simple setting in which $X$ is a feature vector in finite-dimensional Euclidean space
and $Y$ is a binary class label in $\{0,1\}$. 
We assume that $X$ has a density,
and denote its marginal and class-conditional densities by $f_X$ and $f_{X|y}$.
We denote the Bayes optimal classifier 
$g^*(\cdot)$ 
and its associated Bayes optimal probability of misclassification by $L^* = P[g^*(X) \neq Y]$. 

The goal is to learn a classification rule $g_m(\cdot)$ from the training data, which maps feature vectors to class labels
such that the
probability of misclassification
$L(g_m) = P(g_{m}(X) \neq Y)$
is small 
(see \cite{devroye1997probabilistic}, page 2).
We will evaluate performance via the
expectation (taken over the training data) of probability of misclassification,
$\E[L(g_m)]$.
In this setting, choosing $g_m \in \mathcal{C}$ via empirical risk minimization (ERM) yields $\E[L(g_m)] \to L^*$ as $m\to\infty$
(see \cite{devroye1997probabilistic}, Theorem 4.5).

\subsection{A Latent Position Graph} \label{sec:setup}

The generative model of interest in this paper involves
$$(X_1,Y_1), \cdots, (X_n,Y_n) \stackrel{iid}{\sim} F_{XY}$$
with $(X_i,Y_i) \in \Real^d \times \{0,1\}$.
It is required that $X_i^\top X_j \in [0,1]$, i.e., the inner product is a valid probability.
Rather than observing the $X_i$'s,
we observe a latent position graph $G=(V,E)$ on $n$ nodes with (binary, symmetric, hollow) adjacency matrix $A$,
that is generated by
$$A_{ij} \stackrel{ind}{\sim} Bernoulli(X_i^\top X_j)$$ for $i < j$. 
This is the so-called random dot product graph (RDPG);
see \cite{JMLR:v18:17-448} 
for a recent survey.

For node classification, we observe the $n \times n$ matrix $A$,
and the class labels $Y_i$ are observed for only $m$ nodes with $m < n$.
The task is to classify the remaining $n-m$ nodes.
This is a simple case of the problem studied in \cite{tang2013}.

\subsection{Graph Embedding} \label{sec:graph_emb}

Given the $n \times n$ matrix $A$ and $\{Y_1,\cdots,Y_m\}$,
we approach the node classification task via unsupervised graph embedding followed by subsequent linear classification.
Specifically, we consider embedding, denoted $h: A_{n \times n} \to (\Real^{d'})^n$, via two approaches: the adjacency spectral embedding (the scaled leading eigenvectors \textbf{$U_{d'} |\Sigma_{d'}|^{1/2}$} of $A$)
\cite{sussman2012}, and the graph convolutional network (a stack of layers, each of which consists of a graph convolution followed by a pointwise nonlinearity; see Section \ref{sec:GCN} for details) \cite{kipf2016semi,kipf2016variational}. 
Other graph embedding methods, not germane to our investigation but popular nonetheless, include
Laplacian spectral embedding \cite{von2007tutorial},
node2vec \cite{grover2016node2vec}, non-negative matrix factorization \cite{lee1999learning, wang2020robust}, etc. See \cite{hamilton2021book} for an overview.

For either ASE or unsupervised GCN,
we denote $\hat{\bm{X}} = \{\hat{X}_1,\cdots,\hat{X}_n\}$ as the embedding.
A classifier $\hat{g}(\cdot)$ is trained on 
$\{(\hat{X}_i,Y_i)\}_{i \in \{1,\cdots,m\}}$.
For $i>m$ the classified label is $\hat{g}(\hat{X}_i)$, and performance is measured via $\E[L(\hat{g})]$. Note that $d'$ is the embedding dimension we choose (e.g., the embedding dimension in ASE, or the neuron size of the last layer in GCN), while $d$ as in Section 2.2 is the true latent position dimension.

Here $\E[L(g_m)]$ is the target performance -- the performance we would achieve if we observed the latent positions themselves.
We cannot hope to do better,
via an argument reminiscent of the
data processing lemma,
as the inner products $X_i^\top X_j$ are corrupted through the Bernoulli noise channel into $A_{ij}$.
RDPG theory for ASE into dimension $d'=rank(\E[A])$
(specifically the $2\to\infty$ norm convergence result presented in
\cite{cape2019})
implies that 
$\E[L(\hat{g})] \to \E[L(g_m)]$ as $n\to\infty$. In other words, ASE is consistent for the classification task. 

However, to the best of our knowledge, there is no such guarantee of consistency for GCN despite its empirical success in 
applications. Theoretical work for GCN has been focused on its expressive power \cite{scarselli2009gnn, xu2019powerful, morris2019higher, chen2019equivalence, chen2020graph, du2019graph, li2018deeper, oono2019graph, Xu2020What}, stability \cite{gama2020stability} and transferability \cite{levie2020transferability, ruiz2020graphon}. As pointed out in \cite{Loukas2020What}, most impossibility results (such as \cite{xu2019powerful, morris2019higher, chen2020graph}) are based on \textit{anonymous} GCN where nodes have the same attributes. In contrast, \textit{non-anonymous} GCN with discriminative node features are provably more powerful, which could be Turing universal given sufficient capacity and the right learning procedure \cite{Loukas2020What}. Indeed, the learning procedure is crucial, as some graph neural networks generalize better than others \cite{Xu2020What}. Recent studies also suggest that graph neural networks can be simplified as linear models without negatively impacting their performance \cite{wu2019simplifying}. The GCN algorithm studied in this paper can be viewed as \textit{non-anonymous} GCN embeddings trained from two different learning procedures: unsupervised versus semisupervised.  



\section{Failure Geometry\label{sec:FailureGeometry}}

In this section we present a latent position model such that unsupervised GCN fails. We shall start with an auxiliary variable $Z$, then construct the latent position variable $X$ using $srt$ (scale-rotate-translate) transformation. The $srt$ transformation is the key to the desired failure geometry, where the first eigenvector of $A$ is orthogonal to the support of $X$ (i.e., the classification signal). It changes the graph topology but preserves the latent distribution.

\subsection{The $srt$ Transformation}
Given a univariate $Z \in [0,1]$, we define the $srt$ transformation $srt( \cdot): [0,1] \rightarrow \Real^2$:
$$
  X = srt(Z) =
\left[\begin{array}{lr}
cos(r)  & -sin(r) \\
sin(r) & cos(r)
\end{array}\right]
 \left[\begin{array}{c}
sZ \\ 0
\end{array}\right]
+ t,
$$
where $s \in \Real_+$ is the scalar term, $r \in [0,2\pi)$ is the rotation angle, and $t \in \Real^2$ is the translation term. For $X$ to be a valid latent position for RDPG (namely $X_i^\top X_j \in [0,1]$), the range of $(s,r,t)$ needs to be a proper subset of $\Real_+ \times [0,2\pi) \times \Real^2$.

As such,
we may assume without loss of generality that the marginal density $f_X$ is supported on $S \subset \Real^2_+ \cap B(0,1)$ (unit ball of dimension 2).  Note that
$\E[A_{srt}]$ is positive semidefinite
and, except for very special values for $(s,r,t)$ for which $S$ is on a ray from the origin, we have rank$(\E[A_{srt}]) = 2$. Both $L^*$ and $\E[L(g_m)]$ remain unchanged under the $srt$ transformation.

\subsection{Visualize the Latent $X$}
\begin{figure}[h!t]
\begin{center}
\includegraphics[width=8cm]{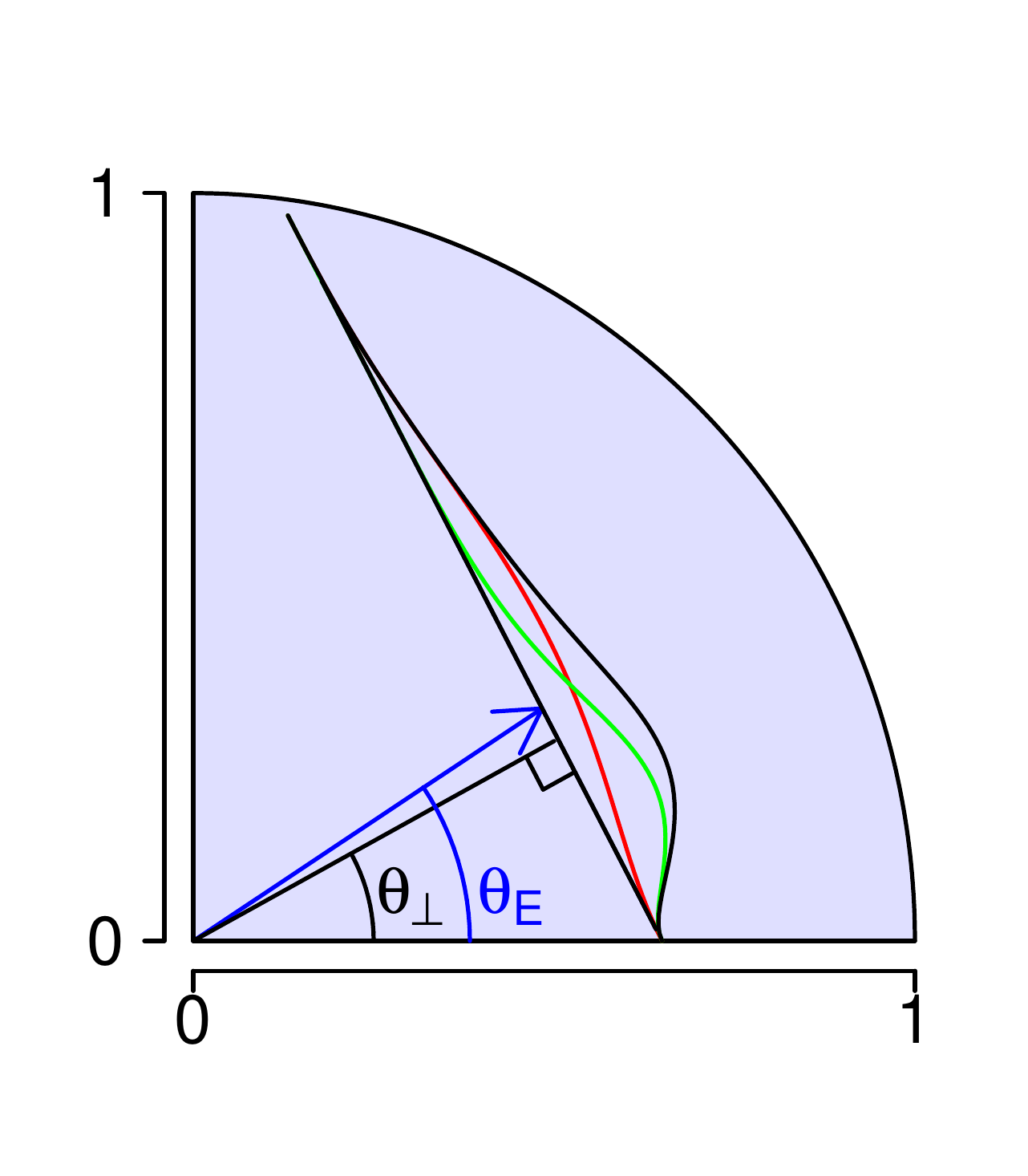}
\includegraphics[width=8cm]{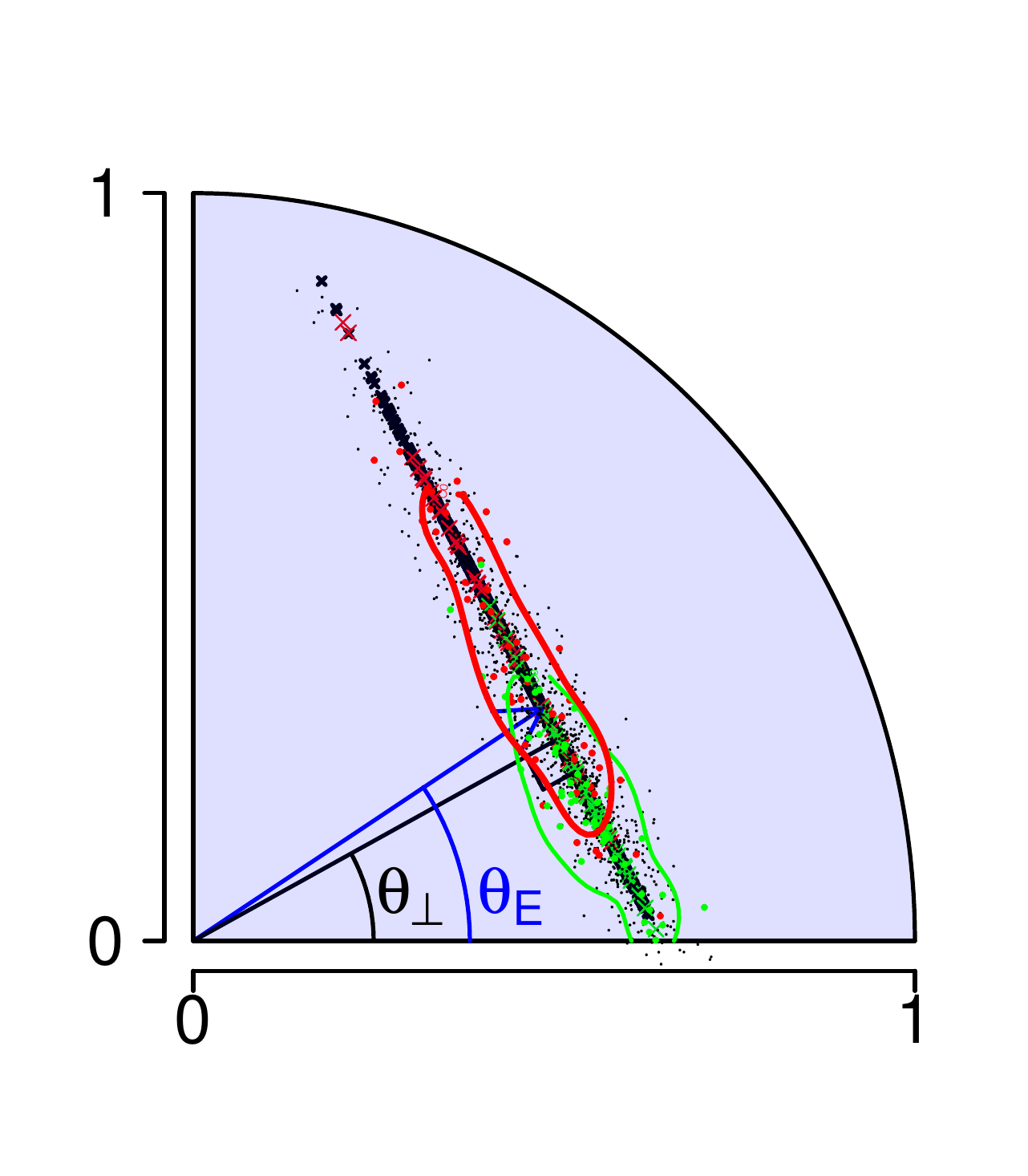}
\end{center}
\caption{Geometry for the canonical case where ASE succeeds but GCN fails. The top panel shows the density: the black line represents density of $f_X$, while the green line and red line represent the class-conditional density of $f_{X|0}$ and $f_{X|1}$. The bottom panel shows one sample data generation with $n=1000$ and $m=100$: the green crosses and red crosses are training points from $f_{X|0}$ and $f_{X|1}$ respectively with known label, while the black crosses are the remaining test points. Round dots are ASE after Procrustes with same color setting. Contours are from kernel density estimation of red dots and green dots. For this case, indicated by the vertical line at $|\theta_E - \theta_\perp| = \pi/32$ in Figure \ref{fig3},
both ASE into dimension $d'=1$ and unsupervised two-layer GCN
perform poorly while ASE into two dimensions has nearly optimal performance.
} 
\label{fig1}
\end{figure}

Figure~\ref{fig1} illustrates the failure mode for GCN.
The density $f_X$ is supported on the line segment $S \subset \Real^2_+ \cap B(0,1)$.
There are two angles of interest:
$\theta_\perp$ is the angle between $\Real_+$ and the ray from the origin perpendicular to $S$;
$\theta_E$ is the angle between $\Real_+$ and the ray from the origin to $\E[X]$.
When $\theta_E = \theta_\perp$, the first eigenvector of $\E[A_{srt}]$ is orthogonal to $S$ and has no classification signal. 

In Figure~\ref{fig1}, we let
$f_{X|0} = srt(Beta(10,3))$, $f_{X|1} = srt(Beta(4,3))$, $(s,r,t)=(1.1,53\pi/32,[0.13,0.97]^\top)$, and class prior probability being $0.5$ (so each class is equally likely).
In this case,
$\theta_\perp = 5\pi/32$ and 
$\theta_E = 3\pi/16$
and hence
$|\theta_E - \theta_\perp| = \pi/32$, which is designed to be almost orthogonal but not exactly.

The $srt$-transformed graph is approximately regular with almost constant node degree. As shown in the simulation section, unsupervised GCN (and also ASE into dimension $d'=1$) is only able to recover the first eigenvector, thus performs poorly for $A_{srt}$. This phenomenon may be generalized to other approximately regular graphs whose inference signal lies on non-leading eigenvectors.

\section{Failure Simulation} \label{sec:simulation}

In this section, we first introduce GCN in both unsupervised and semi-supervised formulations, followed by the node classification results under $A_{srt}$ with varying parameter settings, then provide simulations to demonstrate eigenvector recovery in approximately regular graphs.

\subsection{Graph Convolutional Networks}\label{sec:GCN}

In this section we describe the architecture of unsupervised GCN and semisupervised GCN. The former is directly comparable with ASE as they produce graph embedding without utilizing any additional node information, thus the focus of comparison here. 



We consider the two-layer GCN model \cite{kipf2016semi} designed as
\begin{equation}\label{eq:gcn}
    GCN(\bm{X},A) = \tilde{A} \texttt{ReLU} (\tilde{A} \bm{X} W_0) W_1
    \end{equation}
where $\bm{X}$ is the node feature matrix, $\tilde{A}=D^{-1/2}(A+I)D^{-1/2}$ is the symmetrically normalized adjacency matrix, and $D$ is the node degree matrix. Since our latent position graph does not have node features, the input $\bm{X}$ is set to the $n \times n$ identity matrix by default, which is equivalent to a \textit{non-anonymous} GCN. 
Alternatively, by replacing \texttt{ReLU} activation by the identity activation \cite{wu2019simplifying}, it becomes a linear GCN.

\subsubsection*{Unsupervised GCN}
Unsupervised GCN 
(Variational Graph Auto-Encoder \cite{kipf2016variational})
estimates the latent graph embedding matrix $\hat{\bm{X}}$ using a variational auto-encoder architecture without seeing any labels. The variational inference model of $\hat{\bm{X}}$ is given by
$$q(\hat{\bm{X}} | \bm{X},A) = \prod_{i=1}^n q(\hat{X_i} |\bm{X}, A)$$
with $ q(\hat{X_i} |\bm{X}, A) = \mathcal{N} (\hat{X_i} | \bm{\mu}_i, diag(\bm{\sigma}_i^2)) $.
In our featureless setting where $\mathbf{X} = I$, the parameters of the inference model $\bm{\mu}, \bm{\sigma}$ (the matrix of mean vectors $\bm{\mu}_i$ and variance vectors $\bm{\sigma}_i$ respectively) are estimated by
\begin{align}
    \bm{\mu} &= GCN_{\bm{\mu}} (\bm{X},A) = \tilde{A} \texttt{ReLU} (\tilde{A} W_0) W_1^{\bm{\mu}}  \\
    \log \bm{\sigma} &= GCN_{\bm{\sigma}} (\bm{X},A) = \tilde{A} \texttt{ReLU} (\tilde{A} W_0) W_1^{\bm{\sigma}}  
\end{align}
where $ GCN_{\bm{\mu}}$ and $GCN_{\bm{\sigma}}$ share the first layer weights $W_0$ and learn the second layer weights $W_1^{\bm{\mu}}, W_1^{\bm{\sigma}} $ independently. 

Unsupervised GCN is trained by reconstructing the input graph. It optimizes the variational lower bound $\mathcal{L}$ with respect to $W_0, W_1^{\bm{\mu}}, W_1^{\bm{\sigma}}$:
\begin{equation}
  \mathcal{L}_{unsup} = \E_{q(\hat{\bm{X}} | \bm{X},A) } [ \log p (A | \hat{\bm{X}}) ] - KL[q(\hat{\bm{X}} |\bm{X}, A )|| p (\hat{\bm{X}})]
\end{equation}
where the $KL[q(\cdot) || p(\cdot)]$ term represents the Kullback-Leibler divergence between $q(\cdot)$ and $p(\cdot)$. We assume a Gaussian prior $p(\hat{\bm{X}}) = \prod_i \mathcal{N} (\hat{X}_i | 0, \mathds{1})$. In the training stage, the embedding is estimated using the reparameterization trick \cite{kingma2013auto} as $\hat{\bm{X}} = \bm{\mu} + \epsilon \bm{\sigma}$. After training, the graph embedding $\hat{\bm{X}} = \bm{\mu}$ is used for node classification, as described in Section \ref{sec:graph_emb}.

\subsubsection*{Semisupervised GCN}
The semisupervised GCN produces the class probabilities as
\begin{equation}
 P = \texttt{softmax} (GCN(\bm{X},A) ),   
\end{equation}
trained with $m$ labels via cross-entropy loss
\begin{equation}
\mathcal{L}_{semisup} = -\sum_{i=1}^m \sum_{k=1}^K Y_{ij} \log P_{ij}
\end{equation}
and evaluated on the remaining $n-m$ labels.
Besides deep learning, other types of semisupervised learning methods exist, typically involved constructing the graph from the input data \cite{yuan2021semi}; and see \cite{zhu2005semi} for a survey.

\subsection{Experiment 1}

\begin{figure}[h!t]
\begin{center}
\includegraphics[width=8.5cm]{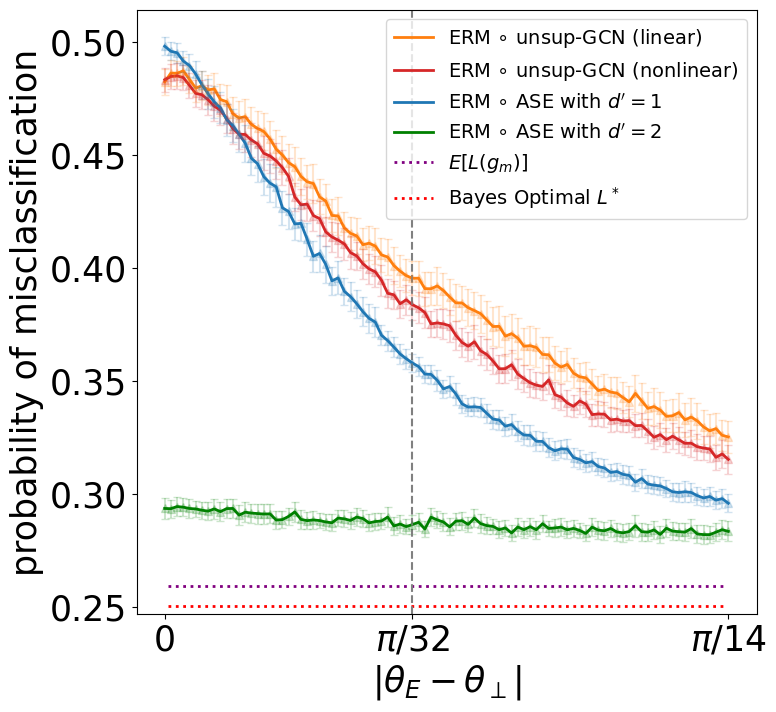}
\end{center}
\caption{Performance for the canonical case where ASE succeeds but two-layer unsupervised GCN fails. ERM in the legend stands for empirical risk minimization classifier. We run $100$ Monte Carlo replicates and report the average classification error.
} 
\label{fig3}
\end{figure}

Here we consider the latent graph $A_{srt}$ as in Section~\ref{sec:FailureGeometry} and Figure~\ref{fig1}. Everything else being the same, we vary the $srt$ transformation as follows: Let $(s,r,t)=(1.1,53\pi/32,t)$, where $t$ ranges linearly from $[0.25,0.97]^\top$ to $[0.02,0.97]^\top$ and preserves the inner product constraint for RDPG. As a result, $|\theta_E - \theta_\perp|$ ranges from $0$ to $\pi/14$, i.e., from exact orthogonality to less orthogonal between the first eigenvector of $E[A_{srt}]$ and the support of $X$. Figure \ref{fig3} shows the classification performance of each method at $m=100$ and $n=1000$ and increasing angle difference. 

For the given $A_{srt}$, the optimal Bayes error is $L^* \approx 0.25$ and $\E[L(g_{100})] \approx 0.26$ regardless of $t$. ASE into two dimensions always works well, while both
unsupervised GCN
and ASE into one dimension
yield much worse performance when $|\theta_E - \theta_\perp| \approx 0$. As the angle becomes less orthogonal, the first eigenvector of $A_{srt}$
captures more classification signal, and all methods perform better. 

Note that nonlinear GCN (Relu activation) performs similarly as linear GCN (linear activation). In this experiment, unsupervised GCN (either linear or nonlinear) uses the two-layer GCN model with hidden size $2$ (because the underlying latent variable lies in $\mathbb R^2$), learning rate $0.01$, no weight decay, and a maximum of $500$ epochs with early stopping at a patience of $100$ epochs. 




\subsection{Experiment 2}

\begin{figure}[h!t]
\begin{center}
\includegraphics[width=9cm]{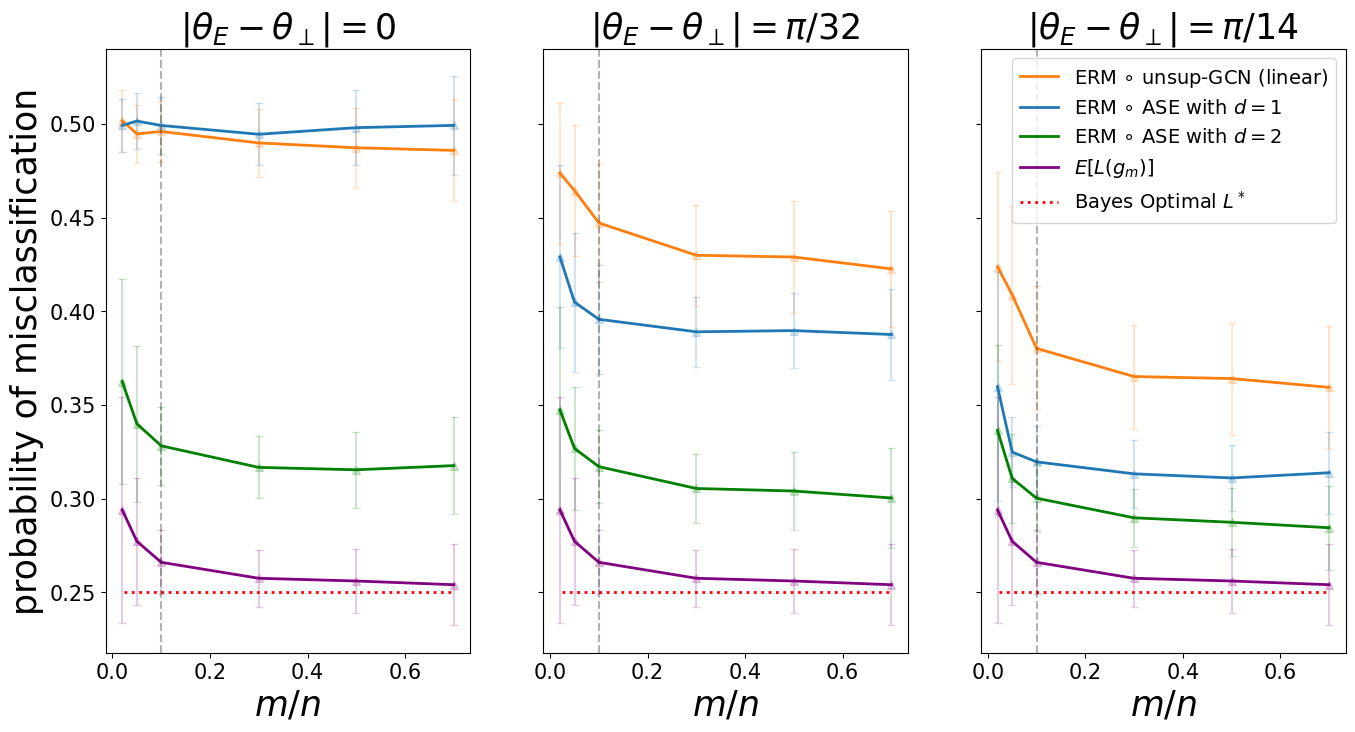}
\end{center}
\caption{Effects of changing $m/n$ on the classification performance for different methods. Vertical dash line indicates $m/n = 0.1$ used in Figure~\ref{fig3}.
} 
\label{fig-mn}
\end{figure}

The same phenomenon always holds regardless of the training data ratio $m/n$. Figure \ref{fig-mn} summarizes the classification error at fixed $n=500$, increasing $m/n$ in $\{0.02,0.05,0.1,0.3,0.5,0.7 \}$, using three different model geometries with $100$ Monte Carlo replicates. As $m$ increases, every method has slightly better classification error, yet ASE at $d=2$ exhibits consistent advantages against unsupervised GCN.

Note that ASE into $d'=2$ always performs well because it selects the correct embedding representation under the latent position model. ASE into $d'=1$ is sub-optimal because it only selects the first eigenvector --- the same reason why unsupervised GCN fails (see Section~\ref{exp4}). As angle increases, ASE with $d'=1$ slightly outperforms unsupervised GCN, because neural networks typically has larger estimation variance. Moreover, Relu activation performs slightly better than linear activation for the same reason: as the graph itself and the first eigenvector are all non-negative in each entry, Relu shall have less estimation variance than linear.


\subsection{Experiment 3}

\begin{figure}[h!t]
\begin{center}
\includegraphics[width=8.5cm]{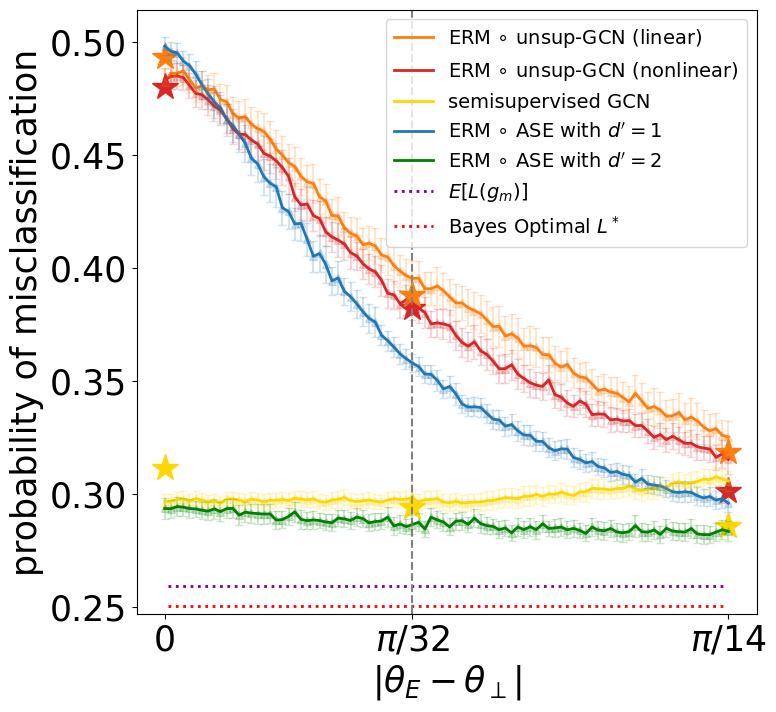}
\end{center}
\caption{Adding parameter cross validation results for unsupervised GCN, and include semisupervised GCN. The curves for GCN models are obtained without hyper-parameter search, while the stars represent the best results after hyper-parameter search. 
} 
\label{punchline2}
\end{figure}

The unsupervised GCN results presented in Section 4.2 are obtained without hyper-parameter search. To show the phenomenon is invariant to parameters, we further present the results using cross validation on the hyperparameters. We also include semisupervised GCN in this experiment, which improves on the unsupervised version but does not outperform ASE with $d'=2$ (again due to the larger estimation variance of neural networks). 

We consider the following parameter in GCN:
\begin{itemize}
    \item Hidden size: \texttt{[2, 4, 8, 16, 32]}
    \item Learning rate: \texttt{[0.01, 0.001]}.
\end{itemize}
The hyper-parameter search is repeated on 3 different weight initializations, and $20\%$ validation set from the training data. We pick the model with the best average training accuracy, and report the testing error.


The results are presented in Figure \ref{punchline2}: there is little change to the performance of unsupervised GCN after optimizing for hyper-parameters (orange stars). The semisupervised GCN performs slightly worse than ASE into two dimensions (yellow line for the no cross validation case, and yellow star for cross validated case). 

\subsection{Experiment 4}
\label{exp4}

In this experiment, we numerically demonstrate that unsupervised GCN only recovers the first eigenvector in certain approximately regular graphs. We consider two settings: the first setting is the same latent position graph $A_{srt}$ as in Figure~\ref{fig1}; then the second setting uses a simple Erdos-Renyi graph where $A_{ij} \stackrel{i.i.d.}{\sim} Bernoulli(0.1)$, for which the degree of each node is about $\frac{n}{10}$. 

In each setting, we compute the unsupervised GCN embedding at output and hidden dimension $4$, then calculate the angle difference between each embedding dimension and each eigenvector of the graph adjacency. The results are summarized in Table 1 for the top three eigenvectors: all GCN dimensions are very close to the leading eigenvector (which is a scalar multiple of ones), and almost exactly orthogonal to the remaining eigenvectors. As ASE at $d'=1$ is the first eigenvector (up-to a scalar), and ASE at $d'=2$ includes the first two eigenvectors, Table 1 can also be viewed as the angle difference between unsupervised GCN and ASE, i.e., GCN embedding is always very similar to ASE at $d'=1$.

The phenomenon is the same throughout small to large $d$, and the embedding is always orthogonal to all remaining eigenvectors of $A$. This illustrates why unsupervised GCN fails in the proposed latent position model $A_{srt}$, and potentially any approximately regular graph whose inference signal lies beyond the first eigenvector. 

\begin{table}
\renewcommand{\arraystretch}{1.3}
\label{table1}
\caption{Computing the angle difference (rounded to two digits) between unsupervised GCN embedding and the eigenvectors of approximately regular graphs at $n=1000$. GCN embedding dimensions are always very close to the first eigenvector, and almost orthogonal to later eigenvectors.}
\centering
\begin{tabular}{c||c|c|c|c|}
\hline
  \multicolumn{5}{c}{The Latent Position Graph} \\\cline{2-5}
 \hline
GCN & Dim 1 & Dim 2 & Dim 3 & Dim 4 \\
\hline
Eig 1 & 0.98  & 0.98 & 0.99 & 0.98 \\
\hline
Eig 2 & 90  & 90 & 90 &90  \\
\hline
Eig 3 & 90  & 90 & 90 &90  \\
\hline
\multicolumn{5}{c}{An Erdos-Renyi Graph} \\\cline{2-5}
 \hline
GCN & Dim 1 & Dim 2 & Dim 3 & Dim 4 \\
\hline
Eig 1 & 2.8  & 2.8 & 2.9 & 2.8 \\
\hline
Eig 2 & 90  & 90 & 90 &90  \\
\hline
Eig 3 & 90  & 90 & 90 &90  \\
\hline
\end{tabular}
\end{table}

\section{Discussion}

The purpose of this short paper is to characterize
a spectral failure mode for graph convolutional networks in the context of the classical statistical learning framework. The failure mode in a latent position graph comes down to: 1) the classification signal lies mostly beyond the leading eigenvector of the graph adjacency matrix; 2) the graph embedding mainly recovers the leading eigenvector. The fundamental idea and geometry insights are presented in its fullest simplicity. 





The failure mode can be generalized to a wide variety of settings, such as multivariate $X$, more complicated generative model and discriminant boundary, etc. Take multivariate $X$ as an example: one may consider the random variable $X \in \Real^d$ (after $srt$ transformation) supported on $S \subset \Real^d_+ \cap B(0,1)$ (unit ball of dimension $d$). $V_E$ is a ray from the origin to E(Z). When $V_E \perp S$, the $srt$-transformed random dot product graph is still an approximately regular graph with classification signal on the non-leading eigenvectors.

The stochastic block model \cite{HollandEtAl1983} offers another special case of our RDPG followed by $srt$ model,
and from \cite{PRDMMSBM} Lemma 3 we see that our model encompasses the mixed membership stochastic block model,
as well as the degree corrected stochastic block model \cite{ZhaoLevinaZhu2012}.
Furthermore, the failure mode can easily be extended to the case of rank$(E[A]) > 2$ with multiple non-leading eigenvectors, and to weighted and/or directed graphs.


Moreover, the use case considered herein ---
a large number of nodes $n$ but a small number of labeled nodes $m$ --- is relevant in many applications, such as text classification \cite{kipf2016semi}, traffic prediction \cite{rahimi-etal-2018-semi}, molecular property prediction \cite{hao2020asgn}. The failure mode has significant implications in molecular applications, as many molecules are regular graphs, e.g., decaprismane $C_{20}H_{20}$ and dodecahedrane $C_{20}H_{20}$ are examples of $3$-regular graphs, see Figure 3 in \cite{sato2020survey}.

Finally, it is not surprising that there exist specific tasks for which GCN perform worse than other methods --- every embedding method has its own bias, and fails when the bias of the task at hand does not coincide with the method’s bias. For example, the Laplacian embedding and the adjacency embedding are both consistent, but one may work better than another depending on the graph structure \cite{Priebe2019}. Moreover, when spectral embedding provides a good representation and the training sample size is small, then it is likely that the added GCN embedding complexity will prove to be a hindrance in terms of the bias-variance trade-off. As noted in \cite{devroye1997probabilistic}: ``Simple rules survive." This short paper provides the foundation for further study: 1) design GCN architectures and learning procedure that are provably better; 2) investigate more general settings to compare GCN and spectral embedding approaches.

\ifCLASSOPTIONcaptionsoff
  \newpage
\fi



%

\bibliographystyle{IEEEtran}

\bibliography{IEEEabrv,ref}

%

%


\begin{IEEEbiography}
[{\includegraphics[width=1in,height=1.25in,clip,keepaspectratio]{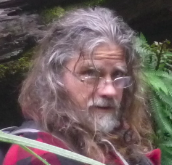}}]{Carey E. Priebe received the BS degree in mathematics from Purdue University in 1984, the MS degree in computer science from San Diego State University in 1988, and the PhD degree in information technology (computational statistics) from George Mason University in 1993. From 1985 to 1994 he worked as a mathematician and scientist in the US Navy research and development laboratory system. Since 1994 he has been a professor in the Department of Applied Mathematics and Statistics at Johns Hopkins University. His research interests include computational statistics, kernel and mixture estimates, statistical pattern recognition, model selection, and statistical inference for high-dimensional and graph data. He is a Senior Member of the IEEE, an Elected Member of the International Statistical Institute, a Fellow of the Institute of Mathematical Statistics, and a Fellow of the American Statistical Association.}
\end{IEEEbiography}
\begin{IEEEbiography}
[{\includegraphics[width=1in,height=1.25in,clip,keepaspectratio]{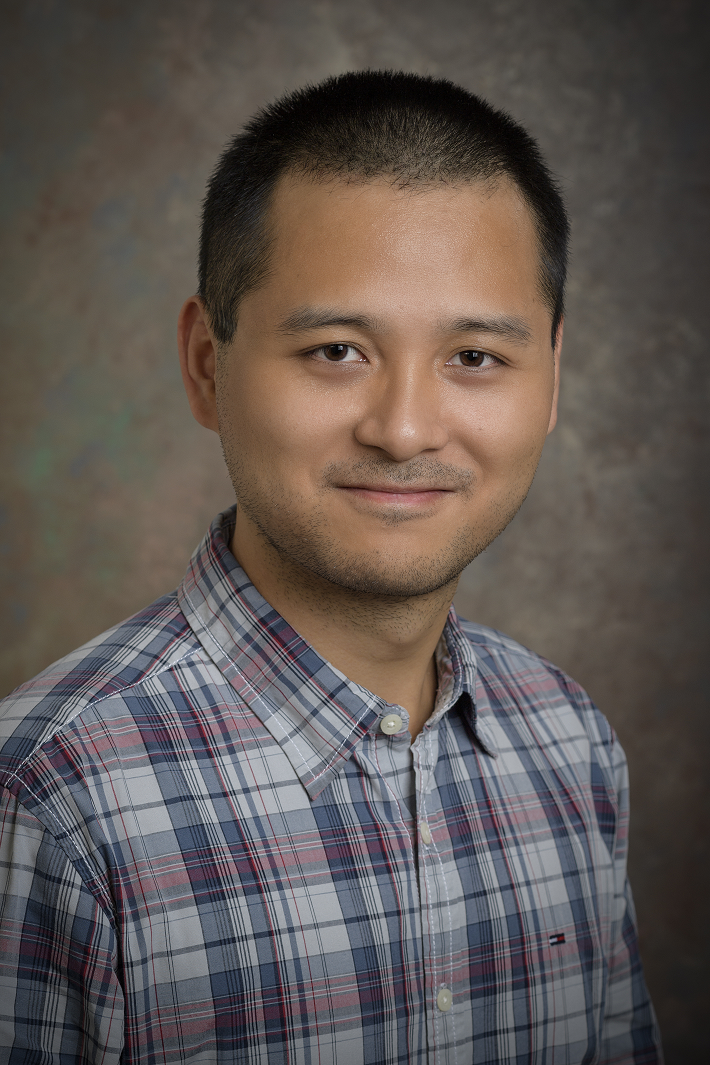}}]{Cencheng Shen received the BS degree in Quantitative Finance from National University of Singapore in 2010, and the PhD degree in Applied Mathematics and Statistics from Johns Hopkins University in 2015. He is 
assistant professor in the Department of Applied Economics and Statistics at University of Delaware. His research interests include testing independence, correlation measures, dimension reduction, and statistical inference for high-dimensional and graph data.}
\end{IEEEbiography}
\begin{IEEEbiography}
[{\includegraphics[width=1in,height=1.25in,clip,keepaspectratio]{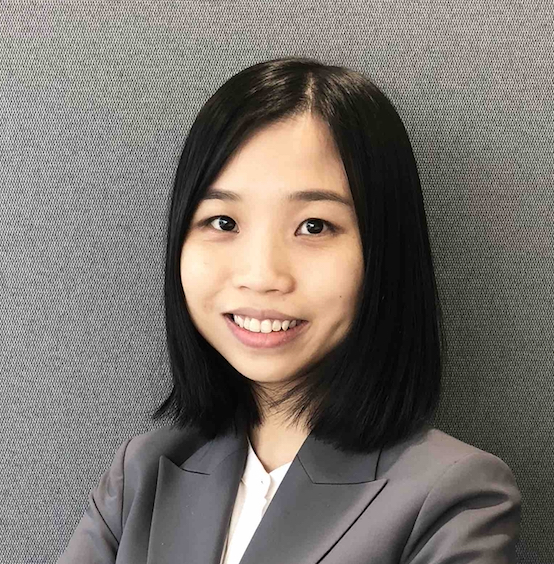}}]{Ningyuan (Teresa) Huang received the BS degree in Statistics from the University of Hong Kong, the MS degree in Data Science from New York University, and is currently pursuing her PhD in the Department of Applied Mathematics and Statistics at Johns Hopkins University. She is interested in representation learning, the theory of deep learning, and inter-disciplinary research in data science.}
\end{IEEEbiography}
\begin{IEEEbiography}
[{\includegraphics[width=1in,height=1.25in,clip,keepaspectratio]{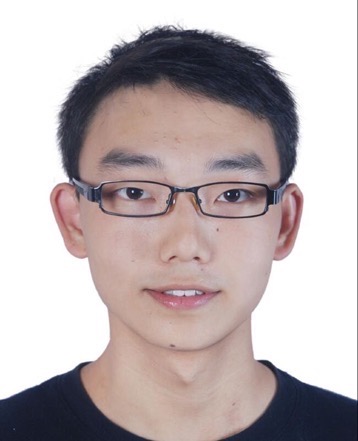}}]{Tianyi Chen received the BS degree in Statistics from Renmin University of China, and is currently pursuing his PhD in the Department of Applied Mathematics and Statistics at Johns Hopkins University.
}
\end{IEEEbiography}







\end{document}